%% file: main.tex
\documentclass[10pt,twocolumn,letterpaper]{article}

 \usepackage{iccv}              % To produce the CAMERA-READY version
\usepackage{times}
\usepackage{epsfig}
\usepackage{graphicx}
\usepackage{subfloat}
\usepackage{amsmath}
\usepackage{amssymb}
\usepackage{multirow}
\usepackage{bbm}
\usepackage{bm}
\usepackage{booktabs}
\usepackage{makecell}
\usepackage{bbding}
\usepackage{subcaption}


\definecolor{iccvblue}{rgb}{0.21,0.49,0.74}
\usepackage[pagebackref,breaklinks,colorlinks,allcolors=iccvblue]{hyperref}

\def\paperID{11833} % *** Enter the Paper ID here
\def\confName{ICCV}
\def\confYear{2025}

\title{PromptMono: Cross Prompting Attention for Self-Supervised Monocular Depth Estimation in Challenging Environments}

\author{Changhao Wang \hspace{10mm} Guanwen Zhang\thanks{Corresponding author} \hspace{10mm} Zhengyun Cheng \hspace{10mm} Wei Zhou\\
 Northwestern Polytechnical University\\
{\tt\small  \{wangch@mail., guanwen.zh@, chengzy@mail., zhouwei@\}nwpu.edu.cn}
}

\begin{document}
\maketitle

\begin{abstract}
Considerable efforts have been made to improve monocular depth estimation under ideal conditions.
However, in challenging environments, monocular depth estimation still faces difficulties.
In this paper, we introduce visual prompt learning for predicting depth across different environments within a unified model, and present a self-supervised learning framework called PromptMono.
It employs a set of learnable parameters as visual prompts to capture domain-specific knowledge.
To integrate prompting information into image representations, a novel gated cross prompting attention (GCPA) module is proposed, which enhances the depth estimation in diverse conditions.
We evaluate the proposed PromptMono on the Oxford Robotcar dataset and the nuScenes dataset. 
Experimental results demonstrate the superior performance of the proposed method.
[Code is being released.]
\end{abstract}

\section{Introduction}
\label{sec:introduction}
Estimating depth from a single RGB image, \emph{i.e.}, monocular depth estimation, is a foundational computer vision task that plays a crucial role in a wide range of applications, including robotics, augmented/virtual reality, and autonomous driving, among others.

\begin{figure}[t]
\begin{center}
\includegraphics[width=80mm]{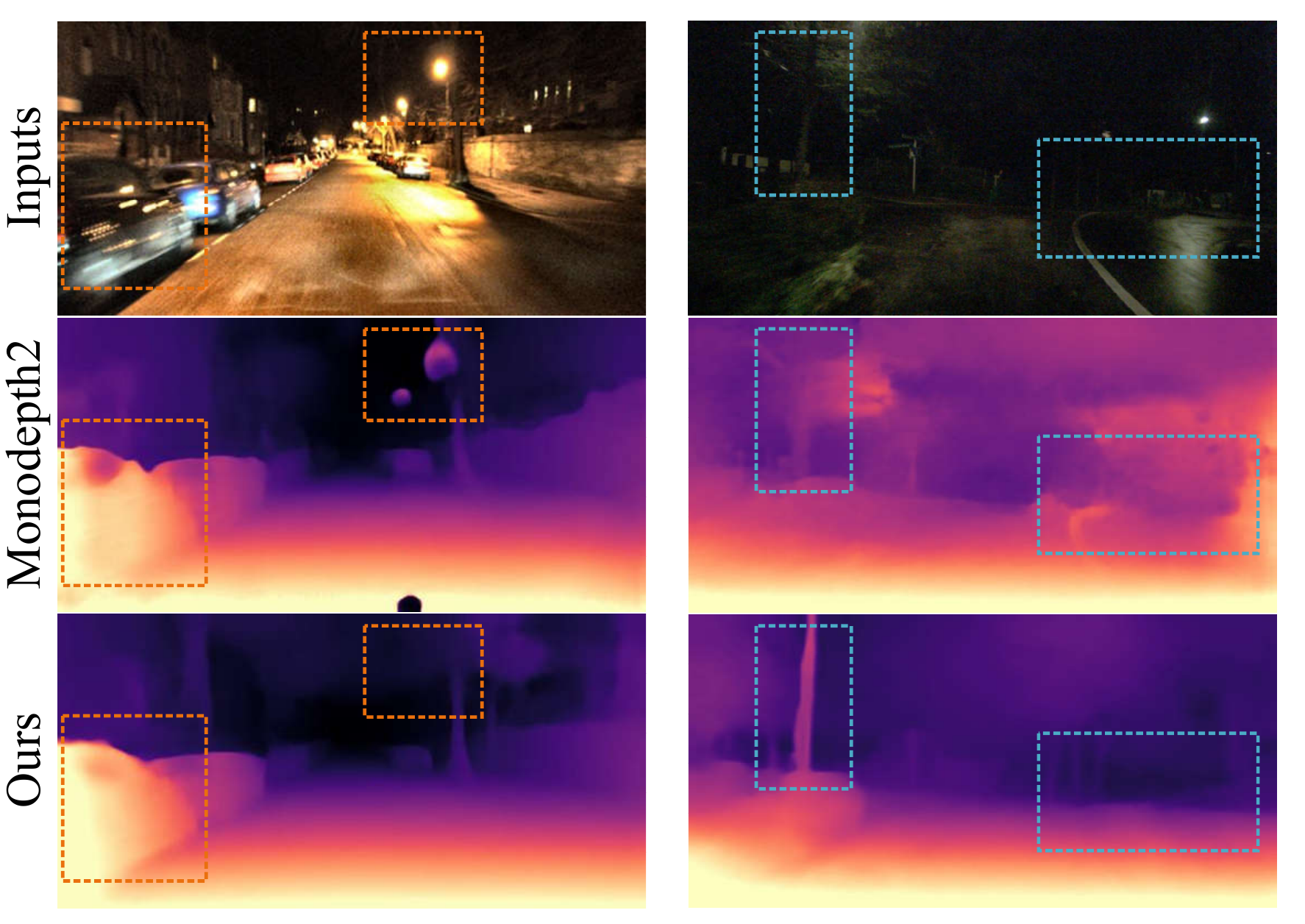}
\end{center}
\caption{
Depth estimation in challenging environments.
The first column and the second column are examples from the Oxford Robotcar dataset~\cite{robotcar} and the nuScenes dataset~\cite{nuscenes}, respectively.
Low visibility at night and reflections on wet roads during rain are detrimental to depth estimation as depicted in the second row.
}
\label{fig:intro}
\end{figure}

Under the assumption of photometric constancy between consecutive images, learning-based monocular depth estimation methods can be conducted in a self-supervised manner, which has garnered significant attention because it eliminates the need for costly 3D ground truth data during the training process~\cite{garg/eccv16, lrc/cvpr17, zhou/cvpr17}.
In recent years, considerable efforts have focused on enhancing loss functions and network architectures for self-supervised monocular depth estimation, resulting in substantial performance improvements~\cite{monodepth/iccv19, hrdepth/aaai21, litemono/cvpr23}.
However, most current research primarily targets depth estimation using images captured under ideal lighting conditions.
In challenging environments, as illustrated in Fig.~\ref{fig:intro}, monocular depth estimation still faces difficulties.
On one hand, differences in image domains mean that directly generalizing daytime depth models to challenging environments can degrade estimation accuracy.
On the other hand, these challenging environments often lead to increased image noise, reduced visibility, and reflections, which can disrupt the photometric constancy assumption and hinder the training of self-supervised monocular depth estimation in such conditions.

Some approaches attempted to address depth estimation during nighttime by enhancing the quality of images captured under poor illumination, ensuring that models could always process data as if under ideal conditions~\cite{adfa/eccv20, rnw/iccv21, steps/icra23}.
However, such methods typically employ specialized image enhancers to process data from a specific domain and find it difficult to be flexibly extended to other domains.
More importantly, enhancing the challenging inputs overlooks the inherent issue of depth models adapting to different environments.
With advancements in the field of image generation, some methods employ generative adversarial network (GAN)~\cite{gan/nips14, cyclegan/iccv17} to create easy-hard image pairs by transforming daytime clear images to challenging samples~\cite{adds/iccv21, itdfa/tetci22}. 
The easy images in these pairs provide high-quality predictions as references for hard images, thereby enhancing the learning of depth estimation in challenging environments.
Building on this foundation, some methods have designed specialized encoder branches to separately extract features from images with varying domains.
Although these specialized branches perform well for their corresponding domains, they do not account for the need of additional networks as the number of domain types increases, which impacts training and inference efficiency.

In this paper, we introduce visual prompt learning to address the challenge of monocular depth estimation across different environments within a unified model and propose a self-supervised learning framework called PromptMono.  
Specifically, PromptMono employs a set of learnable parameters to capture domain-specific knowledge from the data, which serves as visual prompts to enhance the capability of the depth model under diverse domains.
To effectively integrate prompting information into image features, we propose a novel gated cross prompting attention (GCPA) module.
This module first employs a content-gated perception block (CGPB) to process the visual prompts, then leverages 3D depth-wise convolution to project image feature and visual prompts into an embedding space.
Subsequently, the calculation of cross attention between them is performed, thereby generating prompt-enhanced image features for predicting depth.
Besides, we design a self-distillation learning scheme based on image translation to distill the knowledge of the depth model from easy examples to challenging data without requiring a pretrained teacher model.
Evaluation on the Oxford Robotcar dataset~\cite{robotcar} and the nuScenes dataset~\cite{nuscenes} across diverse scenarios demonstrate the superior performance of the proposed PromptMono.

In summary, our major contributions are as follows: 
\begin{itemize}
\item
We present PromptMono, a framework that leverages visual prompt learning with a self-distillation scheme to address the challenge of self-supervised monocular depth estimation in diverse environments.
\item
We propose a novel gated cross prompting attention (GCPA) module that integrates domain-specific information into image representations for predicting depth maps across diverse domains. 
\item
Experimental results demonstrate that the proposed method effectively tackles the challenge of predicting depth across different domains within a unified model and achieves superior performance on the Oxford Robotcar dataset and the nuScenes dataset.
\end{itemize}

\section{Related Work}
\label{sec:related work}
\subsection{Self-Supervised Monocular Depth Estimation}
\label{subsec:ssmde}
Monocular depth estimation is an essential problem and has been studies for years~\cite{eigen/nips14, dorn/cvpr18, yin/iccv19}.
In early, Garg~\emph{et al}.~\cite{garg/eccv16} proposed a view synthesis framework to learn depth by minimizing the photometric error between the reconstructed image pair, which does not need expensive annotated ground truth depth and enlightens the exploration of self-supervised depth estimation.
Zhou~\emph{et al}.~\cite{zhou/cvpr17} proposed a novel method jointly learning depth and relative pose by the photometric loss function in a self-supervised manner.
Godard~\emph{et al}.~\cite{monodepth/iccv19} proposed the per-pixel minimum operation to filter out the occluded and out-of-view pixels that may violate the photometric constancy assumption during the process of view synthesis.
Zhang~\emph{et al}.~\cite{zhang/iros22} introduced the spatio-temporal information to reduce the uncertainty of photometric loss.
Besides, numerous researchers have explored advanced architectures to improve the accuracy of learned depth maps~\cite{hrdepth/aaai21, rmsfm/iccv21, cadepth/3dv21, litemono/cvpr23}.
Lyu~\emph{et al}.~\cite{hrdepth/aaai21} redesign the commonly used skip connection in encoder-decoder architecture and used the nested dense skip connection for depth estimation to learn high-resolution features.
More recently, as the Transformer achieves promising performance in various computer vision fields\cite{transformer/nips17, eccv/CarionMSUKZ20, iccv/StrudelPLS21, vit/iclr21}, some works have introduced attention module into depth networks.
Zhang~\emph{et al}.~\cite{litemono/cvpr23} proposed a novel lightweight backbone called Lite-Mono.
It is constructed based on the combination of convolutional layers and cross-covariance attention modules~\cite{xca/nips21} to capture multi-scale local features and global information, which enables the Lite-Mono to achieve superior accuracy with very lightweight model size.

\subsection{Depth Estimation in Challening Environments}
\label{subsec:dece}
Despite the substantial advances in self-supervised monocular depth estimation, current methods encounter difficulties in performing accurately in challenging environments~\cite{defeat/cvpr20, rnw/iccv21, wsgd/corl22, md4all/iccv23, weatherdepth/icra24}.
Predicting depth in nighttime scenarios is challenging due to the low visibility.
To address this, Spencer \emph{et al}.~\cite{defeat/cvpr20} proposed DeFeat-Net that learns consistent feature representation across domains.
Vankadari \emph{et al}.~\cite{adfa/eccv20} employed a GAN to optimize the depth model, enabling it to extract daytime-like features from nighttime inputs.
Liu \emph{et al}.~\cite{adds/iccv21} used daytime images paired with their corresponding nighttime images transformed by a GAN to train the proposed domain-separated network, which effectively extracts invariant features in nighttime scenarios by eliminating the influence of visual defects.
Zheng \emph{et al}.~\cite{steps/icra23} proposed a nighttime image enhancer that is jointly trained with a depth model.
In addition to nighttime scenarios, depth estimation in adverse weather conditions, such as rainy days, also faces challenges.
Zhao \emph{et al}.~\cite{itdfa/tetci22} employed a daytime model to optimize the domain-specific encoder in each challenging condition, and proposed an approach to evaluate the quality of transformed images for enhancing the learning of depth estimation in challenging scenarios based on image transformation.
Gasperini \emph{et al}.~\cite{md4all/iccv23} proposed a simple learning framework that consistently uses daytime images for view synthesis to compute the photometric loss across all challenging conditions during the self-supervised training process.
This approach effectively improves depth estimation performance in adverse weather conditions without the need for extra domain-specific branches.
 
\subsection{Visual Prompt Learning}
\label{subsec:vpl}
Visual prompt learning, generalized from the field of natural language processing (NLP)~\cite{lb/emnlp21}, has been proven effective in improving the performance of models on downstream tasks by capturing task-relevant information~\cite{vpt/eccv22, zk/ijcv22, zj/cvpr23, promptir/nips23, zm/tcsvt24}.
Jia \emph{et al}.~\cite{vpt/eccv22} employed learnable parameters as task-specific prompts that are updated during downstream tuning while keeping the entire pre-trained large visual model frozen, demonstrating advancements in numerous recognition tasks.
Zhou \emph{et al}.~\cite{zk/ijcv22} proposed a method that models prompt context words with learnable vectors for vision-language models.
Zhu \emph{et al}.~\cite{zj/cvpr23} proposed ViPT, a multi-modal tracking framework based on the modal-relevant prompt learning.
Potlapalli \emph{et al}.~\cite{promptir/nips23} introduced visual prompt learning into the image restoration problem and proposed an all-in-one restoration network named PromptIR.
It uses learnable prompts to capture domain-specific information for dynamically guiding the network without requiring prior knowledge of image degradation.
To our knowledge, this paper may be the first to apply visual prompt learning to solve depth estimation under challenging environments.

\begin{figure*}[t]
\centering
\includegraphics[width=175mm]{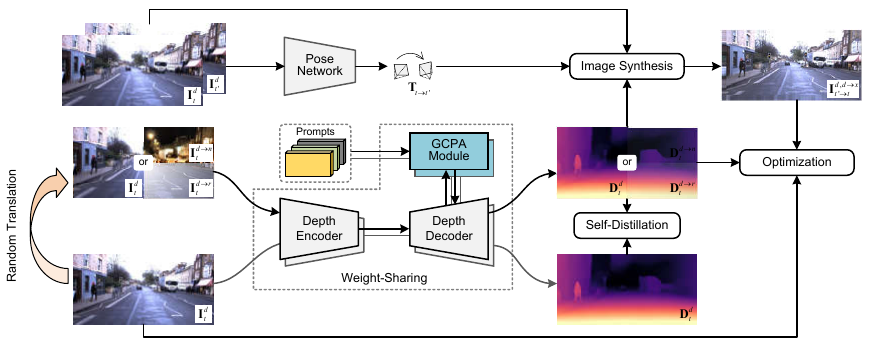}
\caption{
An overview of the proposed PromptMono framework.
It leverages visual prompt learning with a self-distillation scheme to train a unified model for predicting depth in diverse environments.
}
\label{fig:overview}
\end{figure*}

\section{Method}
\label{sec:method}
In this section, we give a brief preliminaries of self-supervised depth estimation.
Following this, we introduce the prompt learning framework and describe the proposed cross prompting attention in detail.
Finally, we present the optimization and inference processes of the proposed method.

\subsection{Preliminaries}
\label{subsec:pre}
Given a target image~$\bm{{\rm I}}_t$ and a source image~$\bm{{\rm I}}_{t'}$, the target depth map~$\bm{{\rm D}}_t$ and the relative pose between the image pair~$\bm{{\rm T}}_{t\!\rightarrow t'}$ can be estimated by a depth network and a pose network, respectively.
Using the estimated depth map, relative pose and known camera intrinsic matrix $\bm{{\rm K}}$, a synthesized target image~$\bm{{\rm I}}_{t'\!\rightarrow t}$ can be sampled from the source image.
\begin{equation}
\begin{aligned}
&\bm{{\rm I}}_{t'\!\rightarrow t}=\bm{{\rm I}}_{t'}\langle proj(\bm{{\rm D}}_t, \bm{{\rm T}}_{t\!\rightarrow t'}, \bm{{\rm K}})\rangle,
\end{aligned}
\label{eq:imgsyn}
\end{equation}
where $\langle \cdot \rangle$ is the sampling operator, $proj$ indicates the pixel coordinates that are reprojected from the target viewpoint into source viewpoint.
Subsequently, the depth network and pose network can be optimized jointly in a self-supervised learning manner by minimizing the photometric loss between the target image and the synthesized one~\cite{lrc/cvpr17}.
Following the widely used framework Monodepth2~\cite{monodepth/iccv19}, the photometric loss is formulated using a per-pixel minimum operation as:
\begin{equation}
\begin{aligned}
L_{\rm p} = \min \limits_{t'} pe(\bm{{\rm I}}_t, \bm{{\rm I}}_{t'\!\rightarrow t}),
\end{aligned}
\label{eq:lp}
\end{equation}
where~$t'\in[t-1, t+1]$ represents the previous and the next frames with respective to the target image, $pe$ denotes the photometric error that is formulated with  SSIM~\cite{ssim/tip04} and L1 loss as:
\begin{equation}
\begin{aligned}
pe(\bm{{\rm I}}_a, \bm{{\rm I}}_b)\!=\!\frac{\alpha}{2} (1\!-\!{\rm SSIM}(\bm{{\rm I}}_a, \bm{{\rm I}}_b))\!+\!(1\!-\!\alpha)||\bm{{\rm I}}_a\!-\!\bm{{\rm I}}_b||_1,
\end{aligned}
\label{eq:pe}
\end{equation}
where~$\alpha$ is a hyperparameter commonly set to $0.85$.

\subsection{Prompting-based Learning of Depth}
\label{subsec:plde}
To address the inherent challenge of adapting depth models to different environments, we propose a prompting-based learning framework using a siamese architecture that incorporates a self-distillation scheme.
An overview of the proposed learning framework is shown in Fig.~\ref{fig:overview}.

\subsubsection{Training Data Transformation}
\label{subsubsec:tdt}
To train the proposed learning framework, image pairs of the same scene under varying conditions are needed.
Following the previous works~\cite{adds/iccv21, md4all/iccv23}, we use the cycleGAN~\cite{cyclegan/iccv17} to transform the daytime images for generating training data.
Given a daytime image $\bm{{\rm I}}^d$, it can be transformed into an image under challenging scenarios $\bm{{\rm I}}^{d\!\rightarrow x}$ using a pre-trained generation model $\mathcal{G}^{d\!\rightarrow x}$.
\begin{equation}
\begin{aligned}
\bm{{\rm I}}^{d\!\rightarrow x} = \mathcal{G}^{d\!\rightarrow x}(\bm{{\rm I}}^d),
\end{aligned}
\label{eq:imgtrans}
\end{equation}
where $x$ indicates one of the challenging conditions, such as nighttime or a rainy day.
For training the depth model under different scenarios, we construct a corresponding image pair $(\bm{{\rm I}}^{h}, \bm{{\rm I}}^{e})$ as input.
Here, $\bm{{\rm I}}^{e}$ indicates an easy sample, such as a daytime image $\bm{{\rm I}}^d$, while $\bm{{\rm I}}^{h}$ represents a corresponding hard sample that is either randomly transformed from the easy sample to challenging conditions or left unchanged, meaning $\bm{{\rm I}}^{h}$ could be $\bm{{\rm I}}^{d\!\rightarrow x}$ or $\bm{{\rm I}}^d$.

\subsubsection{Siamese Architecture}
\label{subsubsec:sa}
As shown in Fig.~\ref{fig:overview}, we use a weight-shared siamese architecture for the depth network during the training process.
The depth network consists of an encoder $\mathcal{E}_{\bm{{\rm D}}}$ and a prompting-based decoder $\mathcal{P}_{\bm{{\rm D}}}$.
Taking a pair of target images $(\bm{{\rm I}}^{h}_{t}, \bm{{\rm I}}^{e}_{t})$ as inputs, the depth encoder extracts feature maps $(\bm{{\rm F}}^{h}_{t}, \bm{{\rm F}}^{e}_{t})$.
\begin{equation}
\left\{\begin{aligned}
&\bm{{\rm F}}^{h}_{t}\!&\!=\!&\!&\!&\mathcal{E}_{\bm{{\rm D}}}(\bm{{\rm I}}^{h}_{t})\\ 
&\bm{{\rm F}}^{e}_{t}\!&\!=\!&\!&\!&\mathcal{E}_{\bm{{\rm D}}}(\bm{{\rm I}}^{e}_{t})
\end{aligned}\right..
\label{eq:encoder}
\end{equation}
Subsequently, the proposed prompting-based depth decoder, which will be detailed in Sec.~\ref{subsec:cpad}, takes the feature maps $(\bm{{\rm F}}^{h}_{t}, \bm{{\rm F}}^{e}_{t})$ and learnable prompts $\bm{{\rm P}}$ as inputs to predict target depth maps $(\bm{{\rm D}}^{h}_{t}, \bm{{\rm D}}^{e}_{t})$ for hard and easy samples, respectively.
\begin{equation}
\left\{\begin{aligned}
&\bm{{\rm D}}^{h}_{t}\!&\!=\!&\!&\!&\mathcal{P}_{\bm{{\rm D}}}(\bm{{\rm F}}^{h}_{t}, \bm{{\rm P}})\\
&\bm{{\rm D}}^{e}_{t}\!&\!=\!&\!&\!&\mathcal{P}_{\bm{{\rm D}}}(\bm{{\rm F}}^{e}_{t}, \bm{{\rm P}})
\end{aligned}\right..
\label{eq:decoder}
\end{equation}

Meanwhile, the pose network takes the target and source easy samples as inputs to predict the relative pose $\bm{{\rm T}}_{t\!\rightarrow t'}$ between the target viewpoint and the source viewpoint.
The previous work md4all~\cite{md4all/iccv23} has shown that learning from easy samples is a simple and effective approach to train the depth network in challenging conditions.
Therefore, we reconstruct a synthesized image as:
\begin{equation}
\begin{aligned}
&\bm{{\rm I}}^{e, h}_{t'\!\rightarrow t}\!&\!=\!&\!&\!&\bm{{\rm I}}^e_{t'}\langle proj(\bm{{\rm D}}^{h}_t, \bm{{\rm T}}_{t\!\rightarrow t'}, \bm{{\rm K}})\rangle,
\end{aligned}
\label{eq:imgsyn2}
\end{equation}
where $\bm{{\rm I}}^{e, h}_{t'\!\rightarrow t}$ indicates sampling from the easy sample based on the predicted depth map of the hard sample.
We then compute the photometric loss of Eq.~\ref{eq:lp} as:
\begin{equation}
\begin{aligned}
L_{\rm p} = \min \limits_{t'} pe(\bm{{\rm I}}^e_t, \bm{{\rm I}}^{e, h}_{t'\!\rightarrow t}).
\end{aligned}
\label{eq:lp2}
\end{equation}

The depth prediction of easy branch in the siamese architecture is only used for providing a reference for the hard sample learning.
We approximate the inverse depth of the hard sample to that of the easy sample, and introduce a self-distillation loss as:
\begin{equation}
\begin{aligned}
L_{\rm sd} = \left\Vert\,\frac{1}{\bm{{\rm D}}^{h}_t}-\frac{1}{\bm{{\rm D}}^{e}_t}\,\right\Vert_2,\ {\rm if}\ h\ {\rm is}\ d\!\rightarrow\!x.
\end{aligned}
\label{eq:lsd}
\end{equation}
By optimizing the loss function, the knowledge of predicting depth in ideal condition from the depth network itself could be distilled to challenging conditions without requiring a pretrained teacher model.  
Notably, the easy branch of our learning framework does not require gradient computation during the training process.

\begin{figure*}[t]
\centering
\includegraphics[width=175mm]{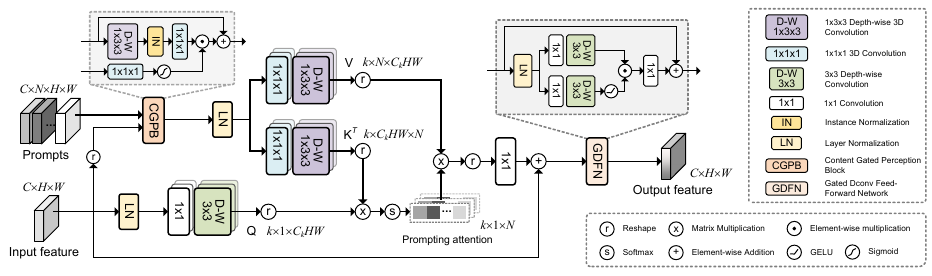}
\caption{
An overview of the proposed GCPA module. 
Taking image feature and visual prompts as inputs, the GCPA module first processes the prompts using the proposed CGPB, and then captures domain-specific information by computing the multi-head cross-attention between the input features and the prompts.
Finally, it outputs the prompt-enhanced features using a gated-DConv feed-forward network~\cite{restormer/cvpr22}.
}
\label{fig:cpa}
\end{figure*}

\subsection{Cross Prompting Attention for Depth}
\label{subsec:cpad}
To effectively incorporate the learnable prompts with the input features across different domains, we propose a gated cross prompting attention (GCPA) module and construct a prompting-based depth decoder.

\subsubsection{Gated Cross Prompting Attention Module}
\label{subsubsec:gcpam}
As demonstrated in Fig.~\ref{fig:cpa}, the GCPA module takes the image feature $\bm{{\rm F}}\in\mathbb{R}^{C\!\times\!H\!\times\!W}$ and the prompts $\bm{{\rm P}}\in\mathbb{R}^{C\!\times\!N\!\times\!H\!\times\!W}$ as inputs to output the prompt-enhanced feature representation.
Here, $C$ is the number of channels, $H\!\times\!W$ is the spatial size, $N$ is the number of prompts.
The prompts are a set of randomly initialized learnable parameters.
Specifically, the GCPA module first employs the proposed content-gated perception block (CGPB) to process the learnable prompts in order to introduce non-linearity.
Then, it generates $query$ ($\bm{{\rm Q}}$) projection from the input feature, while generating $key$ ($\bm{{\rm K}}$) and $value$ ($\bm{{\rm V}}$) projections from the prompts.
The projecting processes are formulated as:
\begin{align}
\bm{{\rm Q}} &= {\rm Linear}_{Q}({\rm LN}(\bm{{\rm F}})),\\
\bm{{\rm K}} &= {\rm Linear}_{K}({\rm LN}({\rm CGPB}(\bm{{\rm P}}, \bm{{\rm F}}))),\\
\bm{{\rm V}} &= {\rm Linear}_{V}({\rm LN}({\rm CGPB}(\bm{{\rm P}}, \bm{{\rm F}}))),
\label{eq:qkv}
\end{align}
where ${\rm Linear}_{Q}$ consists of a $1\!\times\!1$ convolution and a $3\!\times\!3$ depth-wise convolution, while ${\rm Linear}_{K}$ and ${\rm Linear}_{V}$ are consist of a $1\!\times\!1\!\times\!1$ 3D convolution and a $1\!\times\!3\!\times\!3$ depth-wise 3D convolution.
${\rm LN}$ indicates the layer normalization~\cite{LN}.
Here, ${\rm CGPB}$ is defined as:
\begin{equation}
\begin{aligned}
\widetilde{\bm{{\rm P}}}\!=\!{\rm Conv}_{1\!\times\!1\!\times\!1}\!({\rm IN}({\rm Conv}^{\rm D\!-\!W}_{1\!\times\!3\!\times\!3}(\bm{{\rm P}})))\!\cdot\!\sigma({\rm Conv}_{1\!\times\!1\!\times\!1}(\bm{{\rm F}}))\!+\!\bm{{\rm P}}\!,
\end{aligned}
\label{eq:gcpb}
\end{equation}
where ${\rm Conv}_{(\cdot)}$ indicates a convolutional layer, $\sigma$ denotes the sigmoid function, ${\rm IN}$ indicates the instance normalization, ${\rm Conv}^{\rm D\!-\!W}_{(\cdot)}$ indicates a depth-wise convolutional layer.
Next, $\bm{{\rm Q}}$, $\bm{{\rm K}}$, and $\bm{{\rm V}}$ are reshaped and divided into the size of  ${k\!\times\!1\!\times\!C_kHW}$, ${k\!\times\!N\!\times\!C_kHW}$, and ${k\!\times\!N\!\times\!C_kHW}$, where $k$ indicated the number of heads, for calculating the cross prompting attention between the input image feature and the learnable prompts with a multi-head operation.
The calculation is formulated as:
\begin{equation}
\begin{aligned}
\widetilde{\bm{{\rm F}}} = {\rm Conv}_{1\!\times\!1}({\rm Softmax}(\bm{{\rm Q}}\cdot\bm{{\rm K}}^T/ \tau)\cdot\bm{{\rm V}}) + \bm{{\rm F}},
\end{aligned}
\label{eq:cpa}
\end{equation}
where $\tau$ denotes a learnable scaling parameter~\cite{xca/nips21}.
Finaly, the GCPA module generates the prompt-enhanced feature representation as:
\begin{equation}
\begin{aligned}
\hat{\bm{{\rm F}}} = {\rm GDFN}(\widetilde{\bm{{\rm F}}}),
\end{aligned}
\label{eq:ffn}
\end{equation}
where ${\rm GDFN}$ is the gated-Dconv feed-forward network~\cite{restormer/cvpr22} as show in Fig.~\ref{fig:cpa}.

\subsubsection{Prompting-based Depth Decoder}
\label{subsubsec:pdd}
In practice, the depth network is commonly a U-Net architecture, in which the decoder up-samples and skip connects the feature maps from the encoder at multi scales.
The proposed GCPA module can be easily integrated into the decoding process of depth prediction without needing to modify the commonly used decoder architecture.

Let $\{\bm{{\rm F}}_{i}\}_{i=1}^{N}$ denotes the $i$-th feature map generated by the depth encoder from the shallow to deep layers, $\{\bm{{\rm D}}_{s}\}_{s=0}^{S-1}$ denotes the depth map predicted by the decoder at scale of $1/2^s$ relative to the original image size, where $N$ is the number of down-sampling stages, $S$ is the required number of multi-scale predictions.
To construct a prompting-based decoder, we integrate one GCPA module into the deepest level of the decoder, and the computation process is formulated as:
\begin{align}
&\bm{{\rm X}}_{i}\!=\!
\left\{\!
\begin{aligned}
\!&\mathcal{C}([\mathcal{U}({\rm GCPA}(\mathcal{C}(\bm{{\rm F}}_{i+1}), \bm{{\rm P}})), \bm{{\rm F}}_{i}]_{\rm c}), i\!=\!N\!-\!1\\
\!&\mathcal{C}([\mathcal{U}(\mathcal{C}(\bm{{\rm X}}_{i+1})), \bm{{\rm F}}_{i}]_{\rm c}),\ \ \ \ \ \ \ \ \ \ \ \ \ 1\!\leq\!i\!<\!N\!-\!1
\end{aligned}
\right.\!,\\
&\bm{{\rm D}}_s\!=\!
\left\{\!
\begin{aligned}
\!&\mathcal{O}(\bm{{\rm X}}_{s}), &1\!\leq\!s\!<\!S \\
\!&\mathcal{O}(\mathcal{C}(\mathcal{U}(\mathcal{C}(\bm{{\rm X}}_{s+1})))), &s=0
\end{aligned}
\right.\!,
\label{eq:pbdecoder}
\end{align}
where $\bm{{\rm X}}_{i}$ denotes the middle feature of each up-sampling stage in the decoder, $\mathcal{C}$ is a convolutional block consisting of a $3\!\times\!3$ convolutional layer and a non-linear activation function, $[\ \cdot\ ]_{\rm c}$ indicates a channel-wise concatenation operator, $\mathcal{U}$ is a bi-linear up-sampling, $\mathcal{O}$ is the prediction head consisting of a $3\!\times\!3$ convolutional layer and a sigmoid function.

\subsection{Optimization and Inference}
\label{subsec:oi}
In summary, the total loss function used for the optimization of the proposed learning framework is defined as:
\begin{equation}
\begin{aligned}
L = \frac{1}{S}\sum\limits_{s=0}^{S-1}(\lambda_{\rm p}L_{\rm p}^s + \lambda_{\rm sd}L_{\rm sd}^s + \lambda_{\rm smooth}L_{\rm smooth}^s),
\end{aligned}
\label{eq:loss_posedepth}
\end{equation}
where $[\lambda_{\rm p}, \lambda_{\rm sd}, \lambda_{\rm smooth}]$ are the loss weights, $L_{\rm smooth}$ is the edge-aware smoothness loss~\cite{lrc/cvpr17} that is commonly applied to smooth predicted depth map.
\begin{equation}
\begin{aligned}
L_{\rm smooth} = |\partial_x d_t^*|e^{-|\partial_x \bm{{\rm I}}_t|} + |\partial_y d_t^*|e^{-|\partial_y \bm{{\rm I}}_t|},
\end{aligned}
\label{eq:ls}
\end{equation}
where~$d_t^*$ is the mean-normalized inverse depth.

During inference, the depth is predicted under different conditions using a single depth model, without requiring prior domain-specific knowledge for architectural modification or switching.

\section{Experiments}
\label{sec:ecperiments}

\subsection{Datasets}
\label{subsec:datasets}
{\bf Oxford RobotCar Dataset.}
The Oxford Robotcar dataset~\cite{robotcar} is an autonomous driving dataset collected at various times throughout one year in urban scenes.
We use sequences 2014-12-09-13-21-02 and 2014-12-16-18-44-24 for daytime and nighttime depth estimation, respectively.
The left images captured by the Bumblebee XB3 camera in each sequence are used for training and testing.
Following the previous works~\cite{adds/iccv21, rnw/iccv21}, the training data are selected from the first 5 splits of each sequence, and the testing data consist of 451 daytime images and 411 nighttime images that are selected from the other splits.

{\bf nuScenes Dataset.}
The nuScenes dataset~\cite{nuscenes} is a challenging autonomous driving dataset collected in dense traffic situations under diverse scenarios.
It includes not only day and night scenes but also rainy scenes.
Following the md4all~\cite{md4all/iccv23}, we use the official split that contains 15129 images for training and 6019 images for testing.
In the test set, there are 4449 images for day-clear scenes, 602 images for night scenes (including night-rain), and 968 images for day-rain scenes.

\subsection{Implementation details}
\label{subsec:implementation details}
The depth network in PromptMono employs ResNet~\cite{resnet/cvpr16} of different depths as backbones for various experimental setups, which will be illustrated in Sec.~\ref{subsec:results}.
The pose network has the same architecture as the Monodepth2~\cite{monodepth/iccv19}.
The PromptMono is trained using Adam optimizer for 20 epochs with a batch size of 16.
The input images from different scenarios are uniformly distributed within a minibatch.
For Oxford RobotCar dataset, the original images are center-cropped to two different sizes of $1280\times640$ and $1152\times672$, then resized to $512\times256$ and $576\times320$, respectively, as inputs for different settings.
For nuScenes dataset, the input images are resized to~$576\times320$.
The learning rate is set as~$10^{-4}$ initially and reduced to~$10^{-5}$ at the last 5 epochs.
The loss weights are set as~$[\lambda_{\rm p}, \lambda_{\rm sd}, \lambda_{\rm smooth}] = [1.0, 4.0, 0.001]$.
The proposed method is implemented with PyTorch~\cite{Paszke2017AutomaticDI} on a NVIDIA GeForce RTX 4090 GPU.

\subsection{Comparison Results}
\label{subsec:results}
We evaluate the proposed method on the Oxford Robotcar dataset~\cite{robotcar} and the nuScenes dataset~\cite{nuscenes} with the commonly used metrics of absRel, sqRel, RMSE, RMSElog, $\delta < 1.25$, $\delta < 1.25^2$, and $\delta < 1.25^3$~\cite{eigen/nips14}.  

\begin{table*}[t]
\small
\renewcommand\arraystretch{1.1}
\centering
\setlength{\tabcolsep}{1.2mm}{
\begin{tabular}{l|ll|c|c|cccc|ccc}
\hline
\multirow{2}*{Methods}
&\multicolumn{2}{c}{\multirow{2}*{Train data}}\vline
&\multirow{2}*{Backbone}
&\multirow{2}*{\makecell[l]{Max\\Depth}}
&\multicolumn{4}{c}{Error $\downarrow$}
\vline &\multicolumn{3}{c}{Accuracy $\uparrow$}\\ \cline{6-9} \cline{10-12}
& & & & 
&absRel &sqRel &RMSE &{\small RMSElog}
&{\small $\delta\!<\!1.25$} &{\small $\delta\!<\!1.25^2$} &{\small $\delta\!<\!1.25^3$}\\ \hline
\multicolumn{11}{c}{Day} \\ \hline
Monodepth2~\cite{monodepth/iccv19}
                 &M &d &ResNet18 &40m 
                 &0.117 &0.673 &3.747 &0.161 &0.867 &0.973 &0.991\\
Monodepth2~\cite{monodepth/iccv19}
                 &M &n &ResNet18 &40m 
                 &0.306 &2.313 &5.468 &0.325 &0.545 &0.842 &0.937\\
ADDS~\cite{adds/iccv21}
                 &M &dT(n) &ResNet18 &40m 
                 &{0.109} &{0.584} &{3.578} &{0.153}
                 &{0.880} &{\bf 0.976} &{0.992}\\
\bf{Ours}
                &M &dT(n) &ResNet18 &40m  
                &{\bf 0.105} &{\bf 0.570} &{\bf 3.347} &{\bf 0.147}
                &{\bf 0.884} &{\bf 0.976} &{\bf 0.993}\\ \hline
Monodepth2~\cite{monodepth/iccv19}
                 &M &d &ResNet18 &60m 
                 &0.124 &0.931 &5.208 &0.178 &0.844 &0.963 &{0.989}\\
Monodepth2~\cite{monodepth/iccv19}
                 &M &n &ResNet18 &60m 
                 &0.294 &2.533 &7.278 &0.338 &0.541 &0.831 &0.934\\
ADDS~\cite{adds/iccv21}
                 &M &dT(n) &ResNet18 &60m 
                 &{0.115} &{0.794} &{4.855} &{0.168}
                 &{0.863} &{0.967} &{0.989}\\
\bf{Ours}
                &M &dT(n) &ResNet18 &60m 
                &{\bf 0.111} &{\bf 0.784} &{\bf 4.692} &{\bf 0.161}
                &{\bf 0.867} &{\bf 0.968} &{\bf 0.990}\\ \hline
\multicolumn{11}{c}{Night} \\ \hline
Monodepth2~\cite{monodepth/iccv19}
                &M &d &ResNet18 &40m 
                &0.477 &5.389 &9.163 &0.466 &0.351 &0.635 &0.826\\
Monodepth2~\cite{monodepth/iccv19}
                &M &n &ResNet18 &40m 
                &0.661 &25.213 &12.187 &0.553 &0.551 &0.849 &0.914\\
{\small Monodepth2}+{\small cycleGAN}~\cite{cyclegan/iccv17}
                &M &d &ResNet18 &40m 
                &0.246 &2.870 &7.377 &0.289
                &{0.672} &0.890 &0.950\\
ADDS~\cite{adds/iccv21}
                 &M &dT(n) &ResNet18 &40m 
                 &{0.233} &{2.344} &{6.859} &{0.270}
                 &0.631 &{0.908} &{0.962}\\
\bf{Ours}
                 &M &dT(n) &ResNet18 &40m 
                 &{\bf 0.206} &{\bf 2.057} &{\bf 6.497} &{\bf 0.246}
                 &{\bf 0.736} &{\bf 0.917} &{\bf 0.966}\\ \hline

Monodepth2~\cite{monodepth/iccv19}
                 &M &d &ResNet18 &60m 
                 &0.432 &5.366 &11.267 &0.463 &0.361 &0.653 &0.839\\
Monodepth2~\cite{monodepth/iccv19}
                 &M &n &ResNet18 &60m 
                 & 0.580 &21.446 &12.771 &0.521 &0.552 &0.840 &0.920\\
{\small Monodepth2}+{\small cycleGAN}~\cite{cyclegan/iccv17}
                 &M &d &ResNet18 &60m 
                 &0.244 &3.202 &9.427 &0.306 &{0.644} &0.872 &0.946\\
ADDS~\cite{adds/iccv21}
                 &M &dT(n) &ResNet18 &60m 
                 &{0.231} &{2.674} &{8.800} &{0.286}
                 &0.620 &{0.892} &{0.956}\\
\bf{Ours}
                 &M &dT(n) &ResNet18 &60m 
                 &{\bf 0.209} &{\bf 2.474} &{\bf 8.591} &{\bf 0.265} &{\bf 0.704}
                 &{\bf 0.898} &{\bf 0.962}\\ \hline
RNW~\cite{rnw/iccv21}
                 &M* &dn &ResNet50 &60m 
                 &0.185 &1.894 &7.319 &0.246 &0.735 &0.910 &0.965\\
STEPS~\cite{steps/icra23}
                 &M* &dn &ResNet50 &60m 
                 &{\bf 0.170} &{1.686} &{6.797} &{0.234}
                 &{0.758} &{0.923} &{0.968}\\            
\bf{Ours}
                 &M* &dT(n) &ResNet50 &60m 
                 &{0.172}  &{\bf 1.540} &{\bf 6.567} &{\bf 0.233}
                 &{\bf 0.763} &{\bf 0.924} &{\bf 0.972}\\ \hline
\end{tabular}}
\caption{Comparison on the Oxford RobotCar dataset.
M: monocular frames of resolution $512 \times 256$;
M*: monocular frames of resolution $576 \times 320$;
d: day-clear frames;
n: night frames;
T($\cdot$): translated data. 
}
\label{tb:rc}
\end{table*}

\begin{figure}[t]
\centering
\includegraphics[width=80mm]{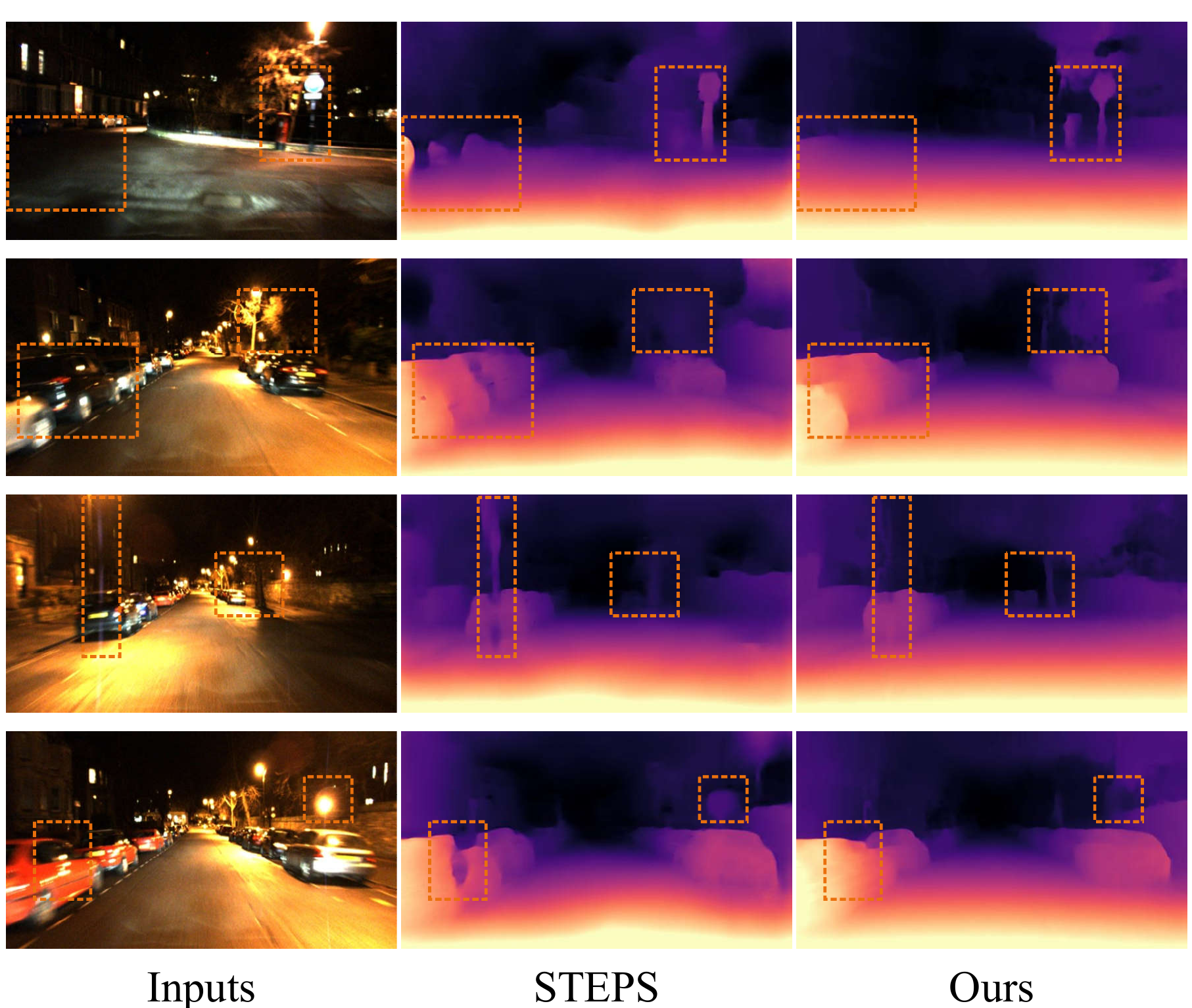}
\caption{
Qualitative results on the Oxford RobotCar dataset.
}
\label{fig:rc}
\end{figure}

\begin{table*}[t]
\small
\renewcommand\arraystretch{1.1}
\centering
\setlength{\tabcolsep}{1.2mm}{
\begin{tabular}{l|ll|ccc|ccc|ccc}
\hline
\multirow{2}*{Methods}
&\multicolumn{2}{c}{\multirow{2}*{Train data}}\vline
&\multicolumn{3}{c}{Day-clear} \vline
&\multicolumn{3}{c}{Night} \vline
&\multicolumn{3}{c}{Day-rain} \\
&&&absRel $\!\downarrow$ &RMSE $\!\downarrow$ &$\delta < 1.25\!\uparrow$ 
  &absRel $\!\downarrow$ &RMSE $\!\downarrow$ &$\delta < 1.25\!\uparrow$ 
  &absRel $\!\downarrow$ &RMSE $\!\downarrow$ &$\delta < 1.25\!\uparrow$ \\ \hline
 
Monodepth2~\cite{monodepth/iccv19}
                 &M* &d
                 &{\bf 0.137} &{\bf 6.692} &{0.850}
                 &{0.283} &{9.729} &{0.518}
                 &{0.173} &{7.743} &{0.776}\\
Monodepth2~\cite{monodepth/iccv19}
                 &M* &dnr
                 &0.148 &6.771 &{\bf 0.853}
                 &2.333 &32.940 &0.105
                 &0.411 &9.442 &0.606\\
RNW~\cite{rnw/iccv21}
                 &M* &dn
                 &0.287 &9.185 &0.562
                 &0.333 &10.098 &0.437
                 &0.295 &9.341 &0.571\\
\bf{Ours}
                 &M* &dT(nr)
                 &{0.145} &{6.709} &0.835
                 &{\bf 0.210} &{\bf 8.442} &{\bf 0.690}
                 &{\bf 0.162} &{\bf 7.435} &{\bf 0.787}\\ \hline
PackNet-SfM~\cite{guizilini/cvpr20}
                 &M*v &d
                 &0.157 &7.230 &0.826
                 &0.262 &11.063 &0.566
                 &0.165 &8.288 &0.771\\                
md4all-AD~\cite{md4all/iccv23}
                 &M*v &dT(nr)
                 &0.152 &6.853 &0.831
                 &0.219 &9.003 &0.688
                 &0.160 &7.832 &0.790\\
md4all-DD~\cite{md4all/iccv23}
                 &M*v &dT(nr)
                 &{\bf 0.137} &{\bf 6.452} &{\bf 0.846}
                 &{\bf 0.192} &{8.507} &{\bf 0.711}
                 &{\bf 0.141} &{\bf 7.228} &{\bf 0.810}\\                
\bf{Ours}
                 &M*v &dT(nr)
                 &{0.145} &{6.666} &{\bf 0.846}
                 &{0.194} &{\bf 8.294} &{0.707}
                 &{0.149} &{7.365} &{0.802}\\ \hline               
\end{tabular}}
\caption{Comparison on nuScenes dataset.
M*: monocular frames of resolution $576 \times 320$;
v: weak velocity supervision;
d: day-clear frames;
n: night frames;
r: day-rain frames;
T($\cdot$): translated data. 
}
\label{tb:ns}
\end{table*}

\subsubsection{Oxford RobotCar Results}
\label{subsubsec:oxford results}
We compare our PromptMono with the advanced methods under various settings, ensuring a fair comparison, and the results are reported in Table~\ref{tb:rc}.
Using the most common backbone ResNet18~\cite{resnet/cvpr16}, our method significantly surpasses ADDS~\cite{adds/iccv21} on all evaluation metrics when evaluating the nighttime depth estimation in both the 40m and 60m depth range settings.
Monodepth2 + cycleGAN~\cite{cyclegan/iccv17} indicates that the Monodepth2~\cite{monodepth/iccv19} model is trained on daytime images and translates nighttime images into daytime for testing.
Although such methods can improve performance when testing on nighttime datasets, the depth model itself lacks generalization ability in nighttime scenarios.
When comparing daytime scenarios, our method also achieves superior performance in both the 40m and 60m depth ranges.
When employing the deeper ResNet50~\cite{resnet/cvpr16} as the backbone and conducting experiments in higher resolution settings, our method achieves the best overall performance compared to RNW~\cite{rnw/iccv21} and STEPS~\cite{steps/icra23}.
Importantly, the proposed PromptMono is an all-in-one method, indicating that it does not require prior domain-specific architectural modifications or a specialized image enhancer for testing.

Qualitative results on the Oxford RobotCar dataset~\cite{robotcar} are shown in Fig.~\ref{fig:rc}.
As depicted in the first two rows of Fig.~\ref{fig:rc}, our method can predict more accurate depth of the road and car in the darkness.
In nighttime scenarios, majority methods tend to estimate depths too close to areas affected by lens flares or bright lights, while the proposed method can identify these areas and provide more reasonable estimations, as illustrated in the last two rows of Fig.~\ref{fig:rc}.

\begin{figure}[t]
\centering
\includegraphics[width=80mm]{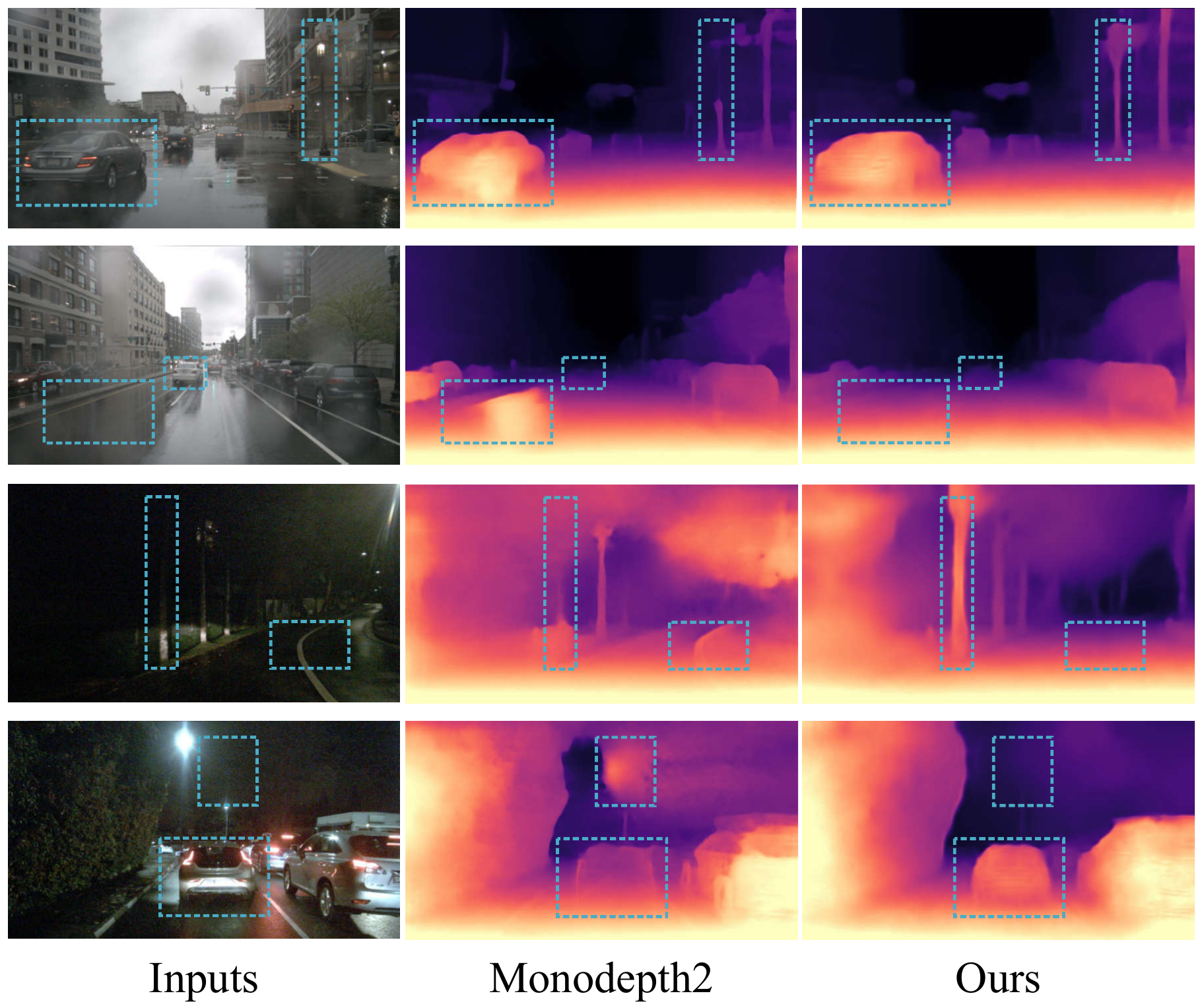}
\caption{
Qualitative results on the nuScenes dataset.
}
\label{fig:ns}
\end{figure}

\subsubsection{nuScenes Results}
\label{subsubsec:nuscenes results}
For comparing on the nuScenes dataset, we use ResNet18~\cite{resnet/cvpr16} as the backbone of the depth network and evaluate the metrics within an 80m depth range.
The comparison results summarized in Table~\ref{tb:ns}.
In fully self-supervised settings, Monodepth2~\cite{monodepth/iccv19}, trained solely on daytime images, shows the lowest estimation errors under day-clear condition, whereas our method performs the best in night and day-rain scenarios.
From PackNet-SfM~\cite{guizilini/cvpr20}, weak velocity supervision is an effective approach for enhancing depth and pose learning by providing absolute scale constraints.
When comparing under weak velocity supervision settings, our method outperforms md4all-AD~\cite{md4all/iccv23} across all scenarios.
md4all-DD~\cite{md4all/iccv23} improves upon md4all-AD by distilling knowledge from a pre-trained daytime depth model, while our proposed method achieves comparable performance in a single-stage training pipeline without requiring a pre-trained teacher model.

We present some qualitative results from the nuScenes dataset~\cite{nuscenes} in Fig.~\ref{fig:ns}.
The first two rows and the last two rows display the results under day-rain and night conditions, respectively.
Reflections on wet roads during rainy days can easily create visual illusions, leading the depth model to make erroneous estimations. 
Our method can mitigate this effect to a certain extent.
Additionally, our method enhances the ability of depth model to perceive better in low visibility environments at night.

\subsection{Ablation study}
\label{subsec:ablation study}

We perform ablation studies on the Oxford Robotcar dataset to demonstrate the effectiveness of the proposed method.
We report the evaluation metrics within a 60m depth range, the details are shown in Table~\ref{tb:ablation} and Table~\ref{tb:ablation2}.

{\bf The benefit of self-distillation learning.}
We use Monodepth2~\cite{monodepth/iccv19}, enhanced with the learning techniques from md4all~\cite{md4all/iccv23}, as our baseline.
From Table~\ref{tb:ablation}, our designed self-distillation learning scheme effectively improves the baseline and also provides external guidance for the prompt learning of depth estimation in challenging environments.

\begin{table}[t]
\renewcommand\arraystretch{1.2}
\small
\centering
\setlength{\tabcolsep}{0.5mm}{
\begin{tabular}{l|cccc|c}
\hline
\makebox[0.1\textwidth][l]{\multirow{2}*{Methods}}
&\multicolumn{4}{c}{Error $\downarrow$} \vline &\multicolumn{1}{c}{Accuracy $\uparrow$} \\ \cline{2-6}
&absRel &aqRel &RMSE &RMSElog &$\delta < 1.25$  \\ \hline
\multicolumn{6}{c}{Day} \\ \hline
Baseline
                &0.116 &0.838 &4.857 &0.170
                &0.864 \\  \hline                     
w/ $L_{\rm sd}$
                &0.111 &0.831 &4.829 &0.166
                &0.864\\
w/ GCPA module       
                &{\bf 0.110} &0.785 &4.841 &0.167
                &{\bf 0.869}\\               
Full model
                &{0.111} &{\bf 0.784} &{\bf 4.692} &{\bf 0.161}
                &{0.867}\\  \hline
\multicolumn{6}{c}{Night} \\ \hline
Baseline
                &0.229 &2.876 &9.037 &0.280
                &0.657\\ \hline                          
w/ $L_{\rm sd}$
                &0.218 &2.678 &8.922 &0.271
                &0.688\\
w/ GCPA module
                &0.214 &2.636 &8.855 &0.276
                &0.690\\
Full model
                &{\bf 0.209} &{\bf 2.474} &{\bf 8.591} &{\bf 0.265}
                &{\bf 0.704}\\  \hline
\end{tabular}}
\caption{Ablation study on the Oxford Robotcar dataset.
}
\label{tb:ablation}
\end{table}

\begin{table}[t]
\renewcommand\arraystretch{1.2}
\small
\centering
\setlength{\tabcolsep}{0.5mm}{
\begin{tabular}{l|cccc|c}
\hline
\multirow{2}*{Prompt Module} 
&\multicolumn{4}{c}{Error $\downarrow$} \vline &\multicolumn{1}{c}{Accuracy $\uparrow$}\\ \cline{2-6}
&{absRel} &{sqRel} &{RMSE} &{RMSElog} &{$\delta < 1.25$}\\ \hline
\multicolumn{6}{c}{Day} \\ \hline
{PGM+PIM}~\cite{promptir/nips23}
               &0.113 &0.831 &4.886 &0.168
               &0.863 \\ \hline
\makecell[l]{\footnotesize GCPA{\footnotesize(w/o CGPB)}}
               &0.112 &0.787 &{\bf 4.821} &{\bf 0.167}
               &{\bf 0.871} \\
GCPA        
                &{\bf 0.110} &{\bf 0.785} &4.841 &{\bf 0.167}
                &0.869 \\ \hline
\multicolumn{6}{c}{Night} \\ \hline
{PGM+PIM}~\cite{promptir/nips23}
               &0.221 &2.830 &9.051 &0.277 
               &0.679\\ \hline
\makecell[l]{\footnotesize GCPA{\footnotesize(w/o CGPB)}} 
               &0.221 &2.730 &8.924 &0.277
               &0.676 \\
GCPA module
               &{\bf 0.214} &{\bf 2.636} &{\bf 8.855} &{\bf 0.276}
               &{\bf 0.690} \\ \hline
\end{tabular}}
\caption{Improvements in visual prompt learning for depth.
}
\label{tb:ablation2}
\end{table}

{\bf The benefit of GCPA module.}
As shown in Table~\ref{tb:ablation}, by integrating the proposed GCPA module into the depth decoder, the depth estimation accuracy ($\delta < 1.25$) gains a slight improvement in daytime scenarios and a significant boost on nightime evaluation.
Meanwhile, GCPA module obviously reduces the estimation error in both conditions.

{\bf Improvements in visual prompt learning.}
To demonstrate the effectiveness of the proposed visual prompt learning in this paper, we conduct more evaluation on the design of GCPA module, as shown in Table~\ref{tb:ablation2}.
Without the proposed content-gated perception block, the performance of depth estimation in nighttime scenarios drops.
Additionally, we employ the prompt learning method proposed in~\cite{promptir/nips23} to train a depth model in the same settings.
Our GCPA module achieves better performance with only an \textbf{additional 0.8M parameters}, while the PGM+PIM requires an additional 1.7M parameters, which shows the proposed cross prompting attention is more effective in prompt learning for depth estimation.

\section{Conclusion}
\label{sec:conclusion}
This paper presents a prompting-based learning framework called PromptMono for self-supervised monocular depth estimation in challenging environments.
A novel GCPA module is proposed to integrate prompting information into image features, thereby enhancing depth estimation across diverse scenarios.
Experimental results across diverse scenarios on the Oxford Robotcar dataset and the nuScenes dataset demonstrate the superiority of the proposed PromptMono in depth estimation. 
Our method is a flexible and general approach that has the potential to be extended to other domains for promising results.

{
    \small
    \bibliographystyle{ieeenat_fullname}
    \bibliography{references}
}

\end{document}